\documentclass{article}
\usepackage{spconf,amsmath,graphicx}
\usepackage{url}
\usepackage{color}
\usepackage{algorithm}
\usepackage{algorithmic}
\usepackage{booktabs} 
\usepackage{hyperref}
\usepackage{subcaption}
\usepackage{multirow}
\usepackage{makecell}
\usepackage{amssymb}

\title{SpMis: An Investigation of Synthetic Spoken Misinformation Detection}

\name{
\begin{tabular}{c}
  Peizhuo Liu$^{1,2,\dagger}$\thanks{$\dagger$ Equal contribution.} \qquad
  Li Wang$^{1,\dagger}$ \qquad
  Renqiang He$^{1,3,\dagger}$ \qquad
  Haorui He$^{1}$ \qquad \\
  Lei Wang$^{4}$ \qquad 
  Huadi Zheng$^{5}$ \qquad 
  Jie Shi$^{4}$ \qquad
  Tong Xiao$^{2}$ \qquad
  Zhizheng Wu$^{1}$ 
\end{tabular}
}

\address{
\begin{tabular}{c}
  $^{1}$ The Chinese University of Hong Kong, Shenzhen, China \\
  $^{2}$ Northeastern University, Shenyang, China \\ 
  $^{3}$ Beijing Institute of Technology, Beijing, China \\ 
  $^{4}$ Huawei International Pte Ltd, Singapore \\ 
  $^{5}$ Huawei Technologies Co., Ltd, China 
\end{tabular}
}

\begin{document}
\ninept
\maketitle
\begin{figure*}[!ht]
    \centering
    \resizebox{1\linewidth}{!}{
    \includegraphics[width=0.8\linewidth]{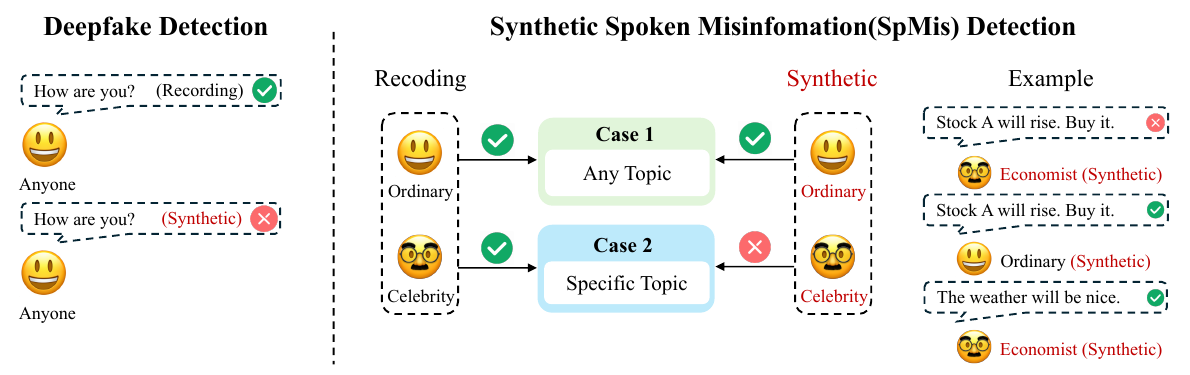}
    }
    \caption{A comparison between DeepFake detection and synthetic spoken misinformation detection. DeepFake detection (left) is to distinguish synthetic and recording. On the other hand, synthetic spoken misinformation detection (right) is to detect synthetic speech by a specific speaker or a group of speaker on specific topics.}
    \label{fig:speaking_info}
\end{figure*}
\begin{abstract}

In recent years, speech generation technology has advanced rapidly, fueled by generative models and large-scale training techniques. While these developments have enabled the production of high-quality synthetic speech, they have also raised concerns about the misuse of this technology, particularly for generating synthetic misinformation. Current research primarily focuses on distinguishing machine-generated speech from human-produced speech, but the more urgent challenge is detecting misinformation within spoken content. This task requires a thorough analysis of factors such as speaker identity, topic, and synthesis. To address this need, we conduct an initial investigation into synthetic spoken misinformation detection by introducing an open-source dataset, SpMis. SpMis includes speech synthesized from over 1,000 speakers across five common topics, utilizing state-of-the-art text-to-speech systems. Although our results show promising detection capabilities, they also reveal substantial challenges for practical implementation, underscoring the importance of ongoing research in this critical area.


\end{abstract}

\section{Introduction}
\label{sec:intro}

People often make significant decisions, such as financial ones, based on information from various sources like news, podcasts, and other media. The spread of misinformation can strongly influence these decisions, leading individuals to make biased choices with serious personal or societal consequences~\cite{barni2022combating}. In the digital age, the accessibility of information through social networks and online platforms has facilitated the rapid and widespread dissemination of misinformation. This misinformation can spread many forms, including text, images, videos and audio content.
In this study, we focus specifically on the phenomenon of synthetic spoken misinformation. As technology advances, the development of sophisticated speech generation techniques has introduced new challenges in the fight against misinformation. Synthetic spoken misinformation, in which advanced speech generation technology is employed to create false or misleading spoken content, presents a unique and growing threat. This type of misinformation often involves the creation of audio recordings that convincingly mimic a particular speaker discussing a specific topic, lending an unwarranted sense of credibility to the false information.

Recently, speech generation has seen significant advancements with the development of various generative models, such as SoundStorm~\cite{borsos2023soundstorm}, VoiceBox~\cite{le2024voicebox}, and NaturalSpeech 3~\cite{ju2024naturalspeech}. By scaling up both datasets and model sizes, zero-shot voice cloning can now produce highly realistic and natural voices using only a few seconds of speech samples from the target speaker \cite{ju2024naturalspeech,amphion,emilia}. While this technology benefits content creators by enabling more engaging productions and offers individuals with speech disabilities a more natural voice~\cite{sharma2012speech}, it also presents the risk of being misused to generate fake information or spread misinformation~\cite{chen2020generalization,mcfend}. There are studies indicating that misinformation spreads faster and wider than non-misinformation~\cite{barni2022combating}. The advancement of speech generation , especially the zero-shot voice cloning techniques, can be misused to create spoken misinformation with minimal cost. It's critical to identify spoken misinformation while prompting the positive use of speech generation.

To address the potential risks of using speech generation technology to create misinformation, the research community has initiated efforts in detecting synthetic speech. The first challenge was organized as the automatic speaker verification anti-spoofing (ASVspoof) in 2015~\cite{wu2015asvspoof,wu2017asvspoof,wu2017asvspoof}, addressing the threat posed by speech generation in the context of automatic speaker verification. This anti-spoofing research has since expanded to include physical attack detection and deepfake detection. However, current speech anti-spoofing methods are predominantly binary classification models that classify speech as either machine-generated or human-produced~\cite{yi2023audio, li2024audio}, as illustrated in Fig.~\ref{fig:speaking_info} (left). While those methods can help prevent the misuse of speech generation technology, they will filter out all the synthetic content even if they are useful for content creation. They don't consider whether the synthetic content contains misinformation or not, instead they filter them out in a brute-force way.

\textit{This study goes beyond the discrimination between synthetic speech and human recordings and focuses on the detection of spoken misinformation}. In other words, we promote the use of speech generation techniques and only detect the synthetic speech that could potentially carry misinformation. The difference between this study and existing deepfake detection is illustrated in Fig.~\ref{fig:speaking_info}. We assume there is a list of celebrity speakers with corresponding list of topics that could potentially carry misinformation if synthetic. The spoken misinformation detection only detects the synthetic speech from the shortlisted speakers and corresponding topics. For the speakers not shortlisted, synthetic spoken misinformation detector will treat them as non-harmful speech, which is usually detected by deepfake detectors.

To our best knowledge, \textit{this is the first work focuses on detecting synthetic spoken misinformation}. This study introduces the first open-source Synthetic \textit{\textbf{Sp}}oken \textit{\textbf{Mis}}information Detection dataset. The identification of synthetic spoken misinformation is based on three key factors: the speaker’s identity, the topic of the speech, and whether the speech is synthesized. The SpMis dataset aims to serve as a comprehensive resource for advancing research in the detection of spoken misinformation. Additionally, we propose a baseline detection system, inspired by Retrieve Augmented Generation, to tackle the challenges posed by misinformation.

\section{Related Work}
\label{sec:relation}

Existing research primarily focuses on the classification of synthetic speech and human recordings based the audio signals. For instance, ASVspoof2015~\cite{wu2015asvspoof} concentrated on detecting spoofed audio generated through voice conversion (VC) and text-to-speech synthesis (TTS). ASVspoof2019~\cite{todisco2019asvspoof} expanded this by using both traditional and state-of-the-art TTS and VC models, incorporating 17 different models to increase the complexity and coverage of spoofed samples. ASVspoof2021~\cite{yamagishi2021asvspoof} sought to evaluate the performance of anti-spoofing systems with more realistic data, including processes like encoding, decoding, and transmission to simulate audio signal transmission over telephone networks. The Audio Deepfake Detection (ADD) challenge, held in 2022~\cite{yi2022add} and 2023~\cite{yi2023add}, introduced new tasks focused on low-quality spoofing, partial spoofing, and spoofing traceback, beyond those covered by ASVspoof. Additionally, the AdvSV dataset~\cite{10446549} was designed for detecting adversarial attacks in audio samples. 

Although the existing studies achieve good performance in distinguishing human recording and synthetic speech from the signal level, they have not examined the content of the synthetic speech, and they detect all the synthetic speech using the same standard (i.e. not distinguishing ``good'' and ``bad'' synthetic content). This work is inspired by the work from the multimodal meme challenge~\cite{kiela2020hateful} that was organized by Meta in 2020 to examine the mismatch between text and images in memes.


\section{SpMis Dataset}
\label{dataset}

This section introduces the design of the synthetic spoken misinformation detection (SpMis) dataset, including the key concepts, general rules and the annotation process of the dataset. The statistics of the dataset is presented in Table~\ref{table:amount} and introduced in detail in this section.

\subsection{Definition of the Synthetic Spoken Misinformation}
\label{sub1}

Synthetic misinformation means that a piece of information is created using synthesis techniques and misleading the general public to make biased decisions. The information that the general public can receive is roughly grouped as the following two scenarios,
\begin{itemize}
    \item \textbf{Case 1:} Any speeches from ordinary people are \textit{not} treated as misinformation, whether the speeches are synthetic or not.
    \item \textbf{Case 2:} A specific topic for recordings of celebrities is fine, while for synthesized celebrities is misinformation.
\end{itemize}
Here, celebrity doesn't mean the celebrities in real world, but to represent the shortlisted identities, while ordinary people mean the non-shortlisted identities.

\subsection{Text Data}
\label{sub2}
We choose five common topics to generate speeches.

\begin{table}[t]
\small
\centering
\caption{Statistics of the SpMis dataset. There are five topics and one other topic. The statistics are presented as numbers of total samples, misinformation samples, speakers and duration for each topic.}
\resizebox{0.9\linewidth}{!}{
\begin{tabular}{lrccr}
\toprule
\textbf{Topic} & \textbf{\# samples} & \textbf{\# misinformation} & \textbf{\# speakers} & \textbf{Duration (hr)} \\
\midrule 
Politics & 76,542 & 1,740 & 772 & 586.59 \\
Medicine & 21,836 & 740 & 1,094 & 429.77 \\ 
Education & 177,392 & 2,970& 989 & 665.59 \\
Laws & 11,422 & 862 & 936 & 1534.78 \\ 
Finance & 53,011 & 2,369 & 940 & 585.69 \\ 
Other & 20,408 & 0 & 1,094 & 1136.23 \\%
\midrule
ALL & 360,611 & 8,681 & 1,094 & 4938.65 \\%
\bottomrule
\end{tabular}
}
\label{table:amount}
\end{table}

\textbf{Finance.} We refer the financial phrase~\cite{malo2014good} as the text data. This work detects the semantic orientations in economic texts and establishes a dataset including annotated financial phrases. The corpus is made out of English news on all listed companies in OMX Helsinki. The news has been downloaded from the LexisNexis database using an automated web scraper. We choose both positive and negative news from the database.

\textbf{Medicine.} Given that in the medical topic, an exact estimate of a disease needs plentiful examination. So regular consults between doctors and patients encompass some irrelevant messages. Therefore we choose a medical abstract dataset~\cite{10.1145/3582768.3582795}. The original corpus contains 28,876 abstracts, which cover neoplasms, digestive system diseases, some general pathological conditions, etc. These abstracts directly describe cases that often happen in the medical topic. We select the training dataset from it, then drop the labels and preserve the plain text.


\textbf{Politics.} Political, especially worldwide, to make the expressions clarified, are conformed to similar habits in speaking. Therefore, we select the dataset of UK parliamentary speeches~\cite{odellevan_2021}, which enjoys decent statements in this area. This dataset ranges from May 1979 to April 2021. Expressions of different periods color the variety. We clean background information and only save a portion of the speech part.

\textbf{Laws.} We choose Super-SCOTUS~\cite{fang2023super} dataset. This corpus connects publicly-available resources including oral arguments and various post-hearing annotations and summaries, including Opinions and case summaries in the Supreme Court of the US(SCOTUS). This provides a comprehensive perspective of researching cases and laws. Besides, this dataset was built to be applied in multiple natural language processing tasks such as classification and prediction. We filter the identity information and save a part of the utterances.

\textbf{Education.} The National Center for Teacher Effectiveness (NCTE) in the US observed 4th and 5th grade elementary mathematics classrooms between 2010 and 2013. The classroom discourses were transcribed as the NCTE Transcripts dataset~\cite{demszky2023ncte}. These transcripts include turn-level annotations for dialogic discourse moves, classroom observation, demographic information, survey responses, and student test scores. These questions and responses illustrate a holistic teaching and learning process. We combined the question-and-answer pairs of teachers and students into complete utterances.



\subsection{Speech Data}
\label{sub3}

The speech generation process needs reference speakers. We choose the Libri-Light~\cite{librilight} dataset as our reference. The audio from it is derived from open-source audio books from the LibriVox project. It is widely used in training speech recognition systems. The data from the Libri-Light dataset is divided into two parts, the limited part with annotated texts and the unlimited part without any texts. There are thousands of speakers from this dataset. The characteristics of these speakers are extracted by our system, and audio is generated using the text data we mentioned above. 

\subsection{Generation Model}
\label{sub4}

As the initial version, we choose two open-source systems for speech data generation.  Amphion~\cite{amphion} and OpenVoice\_v2~\cite{qin2023openvoice} are selected in the initial version of SpMis.

\textbf{Amphion.} The proposed framework encompasses speech generation, music generation, and singing voice conversion. A zero-shot auto-regressive TTS model is trained on Libri-Light, utilizing both texts and corresponding audio. The audio and transcripts in the annotated section of Libri-Light are used for training. For the unannotated section, Whisper-medium\footnote{The model link: \href{https://huggingface.co/openai/whisper-medium}{https://huggingface.co/openai/whisper-medium}\\(released on January 23, 2024, version medium)}~\cite{radford2023robust} is employed for transcription prior to training. The TTS model is built using Llama-style~\cite{touvron2023llama} Transformers with 12 layers, 1024 hidden dimensions, 4096 intermediate hidden dimensions, and 16 attention heads.

\textbf{OpenVoice\_v2.} It offers an efficient method for voice replication using short audio clips. The backbone employs a base TTS model to manage styles and languages, along with a converter to capture the reference speaker's tone color. The exclusion of auto-regressive components accelerates inference. We utilize the pretrained checkpoint\footnote{The checkpoint link: \href{https://github.com/myshell-ai/OpenVoice}{https://github.com/myshell-ai/OpenVoice}\\(released in April, 2024, version V2)}.

\subsection{Generation and Annotation Process}
\label{sub5}

Obviously, taking all the data from above into TTS models can make the dataset redundant and hard to train. We make rules to filter and annotate the data we use.

\textbf{Filtering.} The text datasets encompass a variety of formats and labels tailored for different tasks. We specifically extract paraphrases and dialogues. Due to the imbalance inherent in these datasets, we selectively curate portions from each. Initially, we segment all texts into sentences using full stops. Speech synthesis is performed at the sentence level, with text-specific symbols either removed or substituted with pauses that align with natural speech patterns. Sentences containing fewer than three words are concatenated with adjacent sentences to maintain coherence. For the audio component, to ensure stable generation and fluent output, we select a single audio sample for each reference speaker, with a duration between 5 and 13 seconds. All generated audio is standardized to a sample rate of 16kHz. Given that all generated speech in this context is synthesized, we regard the audio from the Libri-Light dataset as the recording data, representing authentic natural speech. This data is classified as ``other''.

\textbf{Generation.} We get over 1,000 speakers, these speakers are nearly divided half to Amphion, and the other to OpenVoice. To simulate a real scenario, not all speakers are assigned every topic, and the audio length of every assigned topic is not equal either.


\begin{table}[t]
\small
\centering
\caption{The proportion of every part of the annotation. We focus on the \textit{synthetic+celebrity+specific topic} part.}
\resizebox{0.8\linewidth}{!}{
\begin{tabular}{lrr}
\toprule
\textbf{Category} & \textbf{\# samples} & \textbf{Ratio} \\
\midrule 
recordings & 20,408 & 5.66\% \\
\midrule
synthetic$+$ordinary & 305,580 & 84.74\%  \\ 
synthetic$+$celebrity$+$other topics& 25,942 & 7.19\% \\
\textbf{synthetic$+$celebrity$+$specific topic} & 8,681 & 2.41\%  \\  
\bottomrule
\end{tabular}
}
\label{table:data_share}
\end{table}

\textbf{Annotation.} 
The whole dataset is composed of synthetic data and corresponding recording data. For the recording part, we do not conduct extra operations. We use it to do a traditional deepfake detection. This data is ~\textbf{recording} in Table~\ref{table:data_share}. For the synthetic part, given that we get over 1,000 synthetic speakers and corresponding recording speeches, celebrities are few and far between in the whole public, we select 100 speakers with annotated ``celebrity'', which is ~\textbf{synthetic+celebrity+specific topic}. They are randomly specialized in a single topic among the five topics above. Data on topics that are not well versed by them is ~\textbf{synthetic+celebrity+other topics}. The rest part in the synthesized speech is annotated as~\textbf{synthetic+ordinary}. In every topic, we have several items as Table \ref{table:amount} amount displays. These items are excerpts from the text datasets we use above, which range from 10 seconds to minutes. Due to that educational texts are mainly conversation, each conversation round is seen as a single item and the amount of educational topics is significantly more than others.

\section{Detection Methodology}
We design a simple yet effective detection pipeline aiming at misinformation detection. As we mentioned before, we consider detecting in three dimensions. The general detection pipeline is described in Sec \ref{detection}. The involved methods and details are enumerated in Sec \ref{database_found}, and Sec \ref{feature_merger}.

\subsection{Detection Pipeline}
\label{detection}
\begin{figure}[t]
    \centering
     \resizebox{0.9\linewidth}{!}{
    \includegraphics[width=1\linewidth]{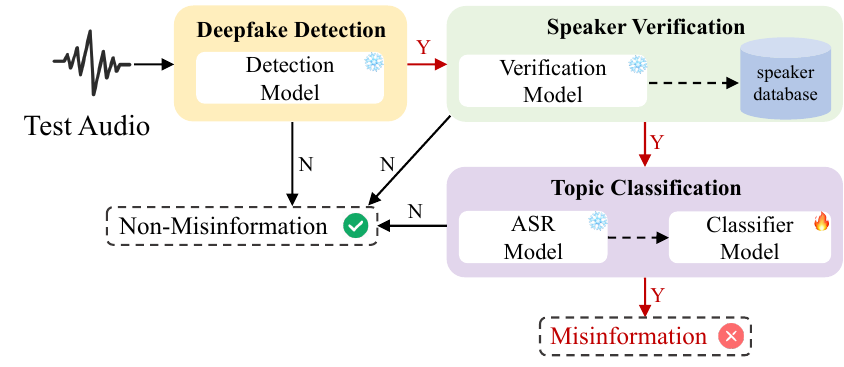}
    }
    \caption{Overview of the detection pipeline. Deepfake Detection checks the synthetic audio and sends it to Speaker Verification. Speaker Verification verifies the celebrities we focus on and sends them to Topic Classification. Topic Classification tells the specific topic. Misinformation is detected through these three modules.}
    \label{fig:pipeline}
\end{figure}

We individually detect the three dimensions as Fig.~\ref{fig:pipeline} shows.

\textbf{Deepfake Detection}. In this module, we employ the AASIST~\cite{Jung2021AASIST} method, which is recognized as an outstanding approach for deepfake detection. During this process, synthetic audio is identified and subsequently directed to the speaker verification procedure for further detection. Conversely, recording audio is disregarded as it is unlikely to contribute to misinformation in our scope.

\textbf{Speaker Verification}. Retrieval-Augmented Generation (RAG) is extensively utilized in the domain of Large Language Models (LLMs). This technique involves storing untrained knowledge embeddings in a database. When an LLM requires information, the relevant knowledge is retrieved from the database and concatenated with the prompts. This approach seamlessly integrates additional knowledge. In our study on misinformation detection, we employ a similar methodology. We utilize the WavLM-SV model, a fine-tuned version of WavLM~\cite{chen2022wavlm}, specifically adapted for speaker verification tasks, as a feature extractor. Features representing the identities of speakers are pre-stored in a vector database. Upon encountering a topic suspected of containing misinformation, the audio is initially processed by WavLM-SV. The extracted representation is then used as a query against the speaker database to retrieve the most similar features. If the similarity exceeds a predefined threshold, the identity of the audio in question is considered to match the retrieved feature, indicating that the synthesized audio includes the speaker of interest, so we assume this audio is from the matched synthetic celebrity. Subsequently, the matched audio proceeds to the next stage of processing, while audio that fails to match is discarded as non-misinformation. Through this approach, once the celebrities we focus on change, the features of corresponding speaker identities can be added or dropped in the database in a plug-in way without training another model.

\textbf{Topic Classification}. In this module, we employ Whisper~\cite{radford2023robust} to transcribe the audio pending detection. Subsequently, the transcriptions are processed by a classifier model for text classification. Upon identifying the specific topic that is not allowed to be said by the specific speaker, we ascertain that the audio has the potential to disseminate misinformation. Conversely, audio that does not match topics is classified as non-misinformation.

\subsection{Speaker Database}
\label{database_found}
We build a speaker database to store the information of shortlisted speakers of interest. Specifically, we use Faiss~\cite{johnson2019billion}, a library developed by Meta for efficient similarity search. Faiss provides various search indices and supports GPU deployment, minimizing frequent I/O operations and significantly accelerating query processing. For every celebrity, to ensure its identity is completely represented, we randomly choose $n$ 10-second clips and throw them into WavLM-SV to get features $f = \{f_1,...,f_n\}$. These features are averaged along the number dimension into a single feature. The features of every speaker are stored as $F_{all}$. This procedure can be concluded as Algorithm \ref{alg1}.

\begin{algorithm}[h]
	\caption{Build a speaker database.} 
	\label{alg1} 
	\begin{algorithmic}[1]
		\REQUIRE $F_{all} \text{ extracted by WavLM-SV, } \text{an empty database } D. $ 
            \FOR{$f$ in $F_{all}$}
		\STATE $f'=\text{mean}(f=\{f_1, ..., f_n\})$ along number dimension
            \STATE $D.\text{add}(f')$ 	
            \ENDFOR
            \RETURN $D$
	\end{algorithmic} 
\end{algorithm}

\begin{table*}[t]
\small
\centering
\caption{Error rates of Speaker Verification module. Ref. Length means the length of reference audio.}
\resizebox{0.85\linewidth}{!}{
\begin{tabular}{lccccccc}
\toprule
\textbf{Ref. Length} & \textbf{Finance(\%)} & \textbf{Laws(\%)} & \textbf{Education(\%)} & \textbf{Politics(\%)} & \textbf{Medicine(\%)} & \textbf{Micro Averaged(\%)} \\
\midrule 
\multirow{1}*{10 seconds}  & 14.82 & 28.07 & 31.18 & 32.24 & 33.11 & 26.78 \\
\multirow{1}*{1 minutes} & 11.90 & 24.36 & 27.98 & 24.54 & 15.95 & 21.52 \\
\multirow{1}*{5 minutes} & 3.84 & 20.19 & 21.82 & 18.16 & 13.78 & 15.33 \\
\multirow{1}*{20 minutes} & 2.11 & 20.19 & 20.51 & 18.56 & 13.51 & 14.47 \\
%
\bottomrule
\end{tabular}}
\label{table:results}
\end{table*}

\subsection{Topic Classification}

To thoroughly inspect specific topics in speech, we implemented a straightforward two-stage approach. First, we employed the state-of-the-art ASR model, Whisper, to transcript the speech. 
Subsequently, we conducted experiments using two different NLP topic topic classification models: BERT~\cite{devlin2019bert} and logistic regression with TF-IDF~\cite{sparck1972statistical} vectorization. BERT excels in topic analysis with its bidirectional context, pre-trained knowledge, and task adaptability. We fine-tuned the BERT model on our dataset, where the textual data extracted by Whisper were tokenized using the BERT tokenizer and then fed into the BERT model. For the logistic regression method, we vectorized the text data using TF-IDF and trained a logistic regression model. The test data were similarly vectorized using TF-IDF, and the trained model was used for prediction and classification performance evaluation.

Given the demonstrated effectiveness of Whisper in speech recognition and the combined use of these models for topic classification, we anticipate our method will yield a high accuracy rate. This two-stage model leverages the strengths of each component, ensuring precise and reliable topic determination.

\begin{table*}[h]
\small
\centering
\caption{Error rates of the Topic Classification module.}
\resizebox{0.9\linewidth}{!}{
\begin{tabular}{lccccccc}
\toprule
\textbf{Model} & \textbf{Train Size} & \textbf{Finance(\%)} & \textbf{Laws(\%)} & \textbf{Education(\%)} & \textbf{Politics(\%)} & \textbf{Medicine(\%)} & \textbf{Micro Averaged(\%)} \\
\midrule 
\multirow{3}*{BERT} & 1,000 & 0.04 & 0.00 & 0.92 & 0.92 & 0.16 & 0.50 \\
~ & 3,000 & 0.04 & 0.00 & 1.13 & 0.07 & 0.16 & 0.40 \\
~ & 10,000 & 0.00 & 0.00 & 0.67 & 0.35 & 0.00 & 0.28 \\
\midrule
\multirow{3}*{Logistic Regression}
& 1,000 & 0.00 & 0.00 & 4.06 & 0.65 & 1.10 & 1.56 \\
~& 3,000 & 0.04 & 0.00 & 1.87 & 0.22 & 0.16 & 0.67\\
~& 10,000 & 0.09 & 0.00 & 1.75 & 0.29 & 0.16 & 0.66\\
%
\bottomrule
\end{tabular}}
\label{table:results_topic}
\end{table*}

\label{feature_merger}

\section{Experiment and Analysis}
In this section, we introduce the hyperparameter settings in Sec \ref{ex_sub1}. The performance of our pipeline and the analysis are in Sec \ref{ex_sub2}. 


\subsection{Experiment Setting}
\label{ex_sub1}
We clarify the data and model setting we use.

\textbf{Model Setting.} For the AASIST model in \textbf{deepfake detection}, we use the default model setting and trained checkpoint\footnote{The configuration link: \href{https://github.com/clovaai/aasist/tree/main/config}{https://github.com/clovaai/aasist/tree/main/config}\\(released on January 18, 2022)}. For the WavLM-SV in \textbf{speaker verification}, we use the frozen fine-tuned checkpoint\footnote{The model link: \href{https://huggingface.co/microsoft/wavlm-base-plus-sv}{https://huggingface.co/microsoft/wavlm-base-plus-sv}\\(released on March 25, 2024)
}. The extracted features $f$ are from 10-second audio, 1-minute audio, 5-minute audio, and 20-minute audio respectively for comparison. 

\begin{figure}[h]
    \centering
     \resizebox{0.85\linewidth}{!}{
    \includegraphics[width=1\linewidth]{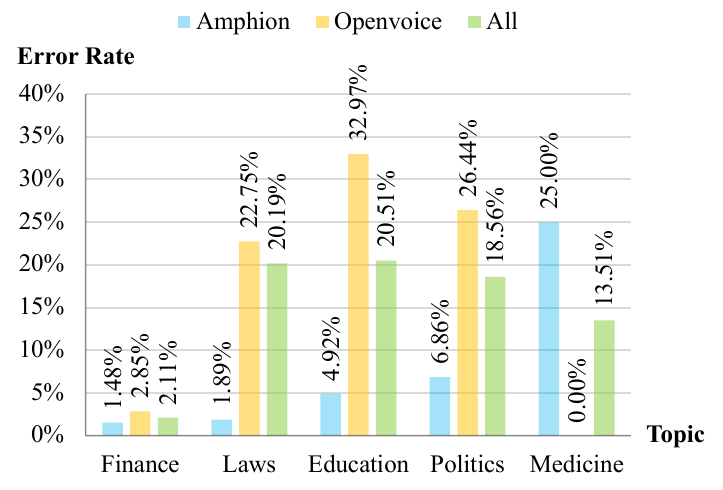}
    }
    \caption{The speaker error rate of two TTS models in Speaker Verification.}
    \label{fig:visual_err}
\end{figure}

Similarity calculation and feature retrieval are based on cosine similarity. The threshold is set to 0.95. The most similar features are retrieved. In \textbf{topic classification}, we also apply the Whisper-medium with default settings and trained checkpoint for transcribing audio. In the text topic extraction experiment, we selected 10,000 samples from the unfiltered dataset as the training set and utilized the filtered dataset as the testing set. This was done to fine-tune the BERT model and train the logistic regression model. For BERT fine-tuning, we use BERT-Base\footnote{The model link: \href{https://huggingface.co/google-bert/bert-base-uncased}{https://huggingface.co/google-bert/bert-base-uncased}\\(released on February 19, 2024)} model and set the training batch size to 8 per device, the learning rate warm-up steps to 500, and the weight decay to 0.01 to prevent overfitting. The evaluation was disabled during training. For the logistic regression model, the maximum number of features for TF-IDF vectorization was set to 5000, and the maximum number of iterations was set to 1000.

\textbf{Data Setting.} All data, encompassing both recordings and our annotated synthetic data, were incorporated into the detection pipeline. Each verification procedure filters the data to be analyzed and excludes non-misinformation data. The outcomes of this process are detailed in Section \ref{ex_sub2}.


\subsection{Results and Analysis}
\label{ex_sub2}
We have different error rates or accuracy in every single detection step. Here, since we concentrate on misinformation detection performance, and as a matter of fact, all the generated speeches hold the same distribution mathematically, we use a dataset that is complete misinformation, namely, our misinformation-annotated dataset (with 6,337 items) to test the performance. All the speakers are celebrities.

\textbf{Deepfake Detection.} Benefiting from its outstanding performance in detecting spoofed speeches within the ASVSpoof dataset, the AASIST model demonstrates significant efficacy in our dataset as well. Notably, without the incorporation of any additional watermarking or anti-spoofing techniques, \textbf{all} synthetic speeches were accurately detected. This outcome reaffirms that the prevailing detection method predominantly focuses on distinguishing between bonafide and spoofed speeches. However, despite its high accuracy, the issue of misinformation remains unaddressed. The synthetic-filtered data is sent to the Speaker Verification module.

\textbf{Speaker Verification.} The data employed in this study is entirely synthetic. Table \ref{table:results} presents the results obtained under different parameters. The 20-minute reference audio captures a greater number of identity characteristics, resulting in a lower error rate compared to the 10-second audio samples. Specifically, the overall error rate (``All'') drops from 26.78\% for 10-second samples to 14.47\%.  Differences were observed across topics during the 20-minute experiment, with the bottleneck might primarily attributed to the model's ability to detect challenging speakers. These variations in speaker detectability across topics resulted in discrepancies in the final accuracy achieved for each topic. However, practical considerations such as time constraints and the availability of lengthy audio samples must be taken into account. The error rate distribution across topics for two TTS models is illustrated in Fig.~\ref{fig:visual_err}. TTS models are potentially deft in specific topics. The significant variability in error rates across different topics underscores the importance of topic-specific tuning and evaluation for misinformation detection. It also highlights that a single detection method may not be universally optimal for all types of content. Future research could explore diverse domains and model configurations, as well as adaptive models that adjust to text complexity in different topics, to further enhance verification accuracy. The speaker-filtered data is sent to the Topic Classification module.

\textbf{Topic Classification.} We randomly selected 1,000, 3,000, and 10,000 pieces of data from the original dataset and input them into Whisper to obtain text as training sets for topic classification. The synthetic-filtered and speaker-filtered data is sent to Whisper for transcription, then for classification. Table~\ref{table:results_topic} presents the results of the text topic classification experiment using BERT and Logistic Regression models across our five different topics. BERT consistently shows lower overall error rates, improving from 0.50\% to 0.28\% as training size increases, while Logistic Regression improves from 1.56\% to 0.66\%. Both models achieve low error rates in Laws, but BERT shows greater improvements in Medicine and Finance. The results suggest that BERT is more effective and reliable for applications requiring high accuracy across diverse topics. 
The data in every check module is recorded. The final mis-detected number and the misinformation detection rate are shown in Table \ref{table:mis_detect}.  This inspires researchers in this topic to pay more attention to the characteristics of audio itself. Besides, the current baseline of misinformation detection reflects the potential for improvement.

\textbf{Unlabeled Data Processing.} 
When an unlabeled sample is introduced into the detection model, the initial step involves evaluating whether the sample is synthesized.  If the sample is identified as synthesized, it is flagged as dubious and sent to the next process.

\begin{table}[h]
\small
\centering
\caption{The misinformation detection result.}
\resizebox{0.9\linewidth}{!}{
\begin{tabular}{lccc}
\toprule
\textbf{Topic} & \textbf{Misinformation} & \textbf{Num. of Errors} & \textbf{Error Rate(\%)} \\
\midrule 
Politics & 1,740 & 333 & 19.14 \\
Medicine & 740 & 102 & 13.78 \\ 
Education & 2,970& 556 & 18.72\\
Laws & 862 & 175 & 20.30 \\ 
Finance & 2,369 & 57 & 2.41 \\ 
\midrule
ALL & 8,681 & 1223 & 14.09 \\%
\bottomrule
\end{tabular}
}
\label{table:mis_detect}
\end{table}

Should the sample pass the deepfake detection, the next phase involves speaker identification. The model attempts to match the speaker’s voice characteristics with those in our existing speaker database. The system retrieves the label of the speaker most similar to the one in the database.  If the similarity score is below the predefined threshold, the system does not consider the person to be in the celebrity database and does not proceed to the next step. The subsequent phase involves conducting a detailed topic analysis of the textual information provided by the speaker. This analysis aims to predict the topic of the speech content accurately.

The final step is to evaluate whether the combination of the predicted speaker label and the predicted topic label exists within a predefined set of valid speaker-topic pairs. This predefined set acts as a benchmark to identify permissible combinations of speakers and topics. If the combination is found within this set, the information is misinformation. Conversely, the inexistence means non-misinformation.


\section{Conclusion and future work}

This study performs an initial investigation of synthetic spoken misinformation detection. A spoken misinformation is defined as a synthetic sample by a specific speaker on a particular topic. To address the concern of the spread of spoken misinformation, we introduced SpMis, the first open-source dataset designed specifically for detecting synthetic spoken misinformation. SpMis encompasses five major topics and includes speech from over 1,000 speakers synthesized using advanced text-to-speech systems. We also propose an intuitive detection approach tailored to this novel task, operating across three dimensions to establish a baseline for future work.

This study is still in the early stage of synthetic spoken misinformation detection. The future efforts will focus on refining the distribution of the SpMis dataset and improving the granularity of data annotations. Additionally, exploring misinformation in other paralinguistic features and developing more sophisticated, potentially end-to-end detection methods will be crucial for enhancing the effectiveness of misinformation detection. We hope our work sparks further research and attention in this vital area, ultimately contributing to stronger defenses against the spread of misinformation through synthetic speech.

\newpage







\bibliographystyle{IEEEbib}
\bibliography{refs}

\end{document}